%% file: root.tex
\documentclass[letterpaper, 10 pt, conference]{ieeeconf}  

\IEEEoverridecommandlockouts

\usepackage{graphics} 
\usepackage{epsfig} 
\usepackage{mathptmx} 
\usepackage{times} 
\usepackage{amsmath} 
\usepackage{amssymb}  

\usepackage{balance}
\usepackage{graphicx}
\usepackage{dsfont}
\usepackage{multicol}
\usepackage{siunitx}
\usepackage[bookmarks=true, hidelinks]{hyperref}
\usepackage{cleveref}
\usepackage{xspace}
\usepackage{xurl}
\usepackage{capt-of}
\usepackage[usenames,dvipsnames,table,xcdraw]{xcolor}
\usepackage{cuted}
\usepackage{titletoc}

\usepackage{booktabs}  
\usepackage{float} 
\usepackage[flushleft]{threeparttable}
\usepackage{amsmath,amssymb,amsfonts}
\usepackage{arydshln} 
\usepackage[ruled,algo2e]{algorithm2e}
\usepackage{mathtools}

\usepackage{wrapfig}
\usepackage{algorithmicx}
\usepackage{algpseudocode}

\usepackage{caption}
\usepackage{stfloats}

\usepackage[
    style=ieee,
    natbib=true,
    citestyle=numeric-comp,
    backend=biber,
    doi=false,
    isbn=false,
    url=false]{biblatex} 
\title{\LARGE \bf
Real-to-Sim Robot Policy Evaluation with \\ Gaussian Splatting Simulation of Soft-Body Interactions
}

\author{Kaifeng Zhang$^{1,2*}$, Shuo Sha$^{1,2*}$, Hanxiao Jiang$^{1}$, Matthew Loper$^{2}$, Hyunjong Song$^{2}$,\\ Guangyan Cai$^{2}$, 
Zhuo Xu$^{3}$, Xiaochen Hu$^{2}$, Changxi Zheng$^{1,2}$, Yunzhu Li$^{1,2}$
\thanks{$^{1}$Columbia University \quad $^{2}$SceniX Inc. \quad $^{3}$Google DeepMind}
\thanks{* Equal contribution. Work partially done while interning at SceniX Inc.}
}

\begin{document}

\maketitle
\thispagestyle{empty}
\pagestyle{empty}

\begin{strip}
    \centering
    \vspace{-30pt}
    \includegraphics[width=\linewidth]{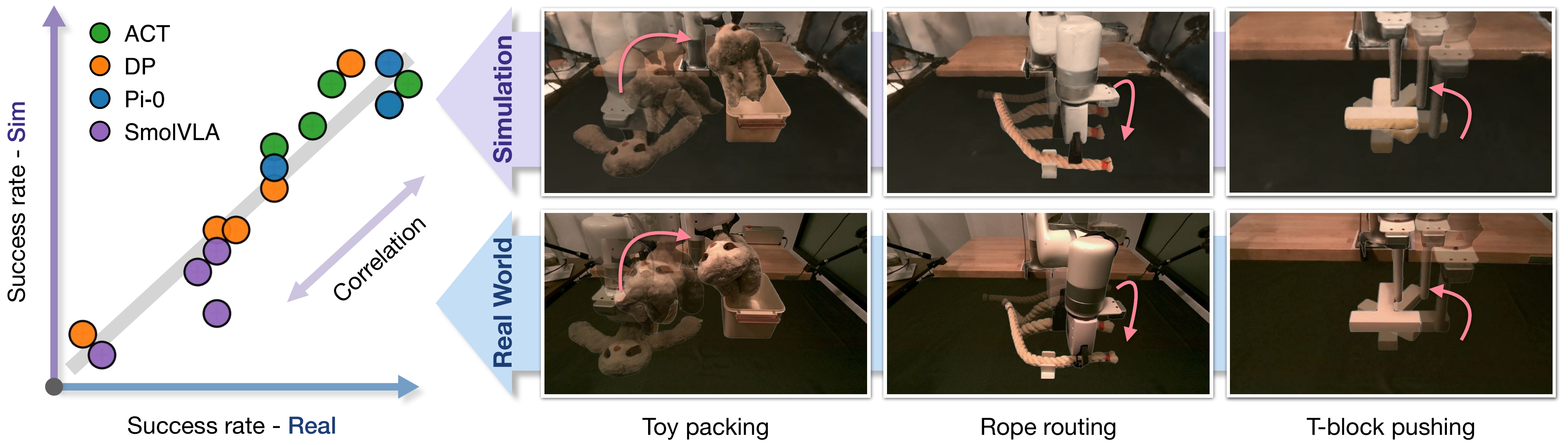}
    \vspace{-15pt}
    \captionof{figure}{\small
    \textbf{Real-to-sim policy evaluation with Gaussian Splatting simulation.}
    \textit{Left:} Correlation between simulated and real-world success rates across multiple policies (ACT~\cite{zhao2023learning}, DP~\cite{chi2023diffusionpolicy}, Pi-0~\cite{black2024pi0}, SmolVLA~\cite{shukor2025smolvla}) shows that our simulation reliably predicts real-world performance.
    \textit{Right:} Representative tasks used for evaluation, including plush toy packing, rope routing, and T-block pushing, are visualized in both real and simulated settings. Our framework reconstructs soft-body digital twins from real-world videos and achieves realistic appearance and motion, enabling scalable and reproducible policy assessment.
    }
    \vspace{-10pt}
    \label{fig:teaser}
\end{strip}

\input{sections/00-abstract}
\input{sections/01-introduction}
\input{sections/02-related-works}
\input{sections/03-method}
\input{sections/04-experiments}
\input{sections/05-conclusion}

\section*{Acknowledgment}

This work is partially supported by the DARPA TIAMAT program (HR0011-24-9-0430), NSF Award \#2409661, Toyota Research Institute (TRI), Sony Group Corporation, Samsung Research America (SRA), Google, Dalus AI, Pickle Robot, and an Amazon Research Award (Fall 2024). This article solely reflects the opinions and conclusions of its authors and should not be interpreted as necessarily representing the official policies, either expressed or implied, of the sponsors.

We would like to thank Wenhao Yu, Chuyuan Fu, Shivansh Patel, Ethan Lipson, Philippe Wu, and all other members of the RoboPIL lab at Columbia University and SceniX Inc. for helpful discussions and assistance throughout the project.

\printbibliography

\clearpage

\begin{center}
\textsc{\Large Appendix}
\end{center}

\newcommand{\DoToC}{
    \hypersetup{linkcolor=black}
  \startcontents
  \printcontents{}{1}{\textbf{Contents}\vskip3pt\hrule\vskip3pt}
  \vskip7pt\hrule\vskip3pt
}

\appendices

\DoToC
\vspace{5pt}

\input{sections/06-appendix}

\end{document}

%% file: sections/00-abstract.tex
\begin{abstract}
    Robotic manipulation policies are advancing rapidly, but their direct evaluation in the real world remains costly, time-consuming, and difficult to reproduce, particularly for tasks involving deformable objects. Simulation provides a scalable and systematic alternative, yet existing simulators often fail to capture the coupled visual and physical complexity of soft-body interactions. We present a real-to-sim policy evaluation framework that constructs soft-body digital twins from real-world videos and renders robots, objects, and environments with photorealistic fidelity using 3D Gaussian Splatting. We validate our approach on representative deformable manipulation tasks, including plush toy packing, rope routing, and T-block pushing, demonstrating that simulated rollouts correlate strongly with real-world execution performance and reveal key behavioral patterns of learned policies. Our results suggest that combining physics-informed reconstruction with high-quality rendering enables reproducible, scalable, and accurate evaluation of robotic manipulation policies. Website: \url{https://real2sim-eval.github.io/}
\end{abstract}

%% file: sections/01-introduction.tex
\section{Introduction}

Robotic manipulation policies have advanced rapidly across a wide range of tasks~\cite{chi2023diffusionpolicy, zhao2023learning, chi2024universal, lin2025sim, tang2023industreal}. However, their evaluation still relies heavily on real-world trials, which are slow, expensive, and difficult to reproduce. As the community shifts toward training foundation models for robotics~\cite{rt22023arxiv, kim2024openvla, black2024pi0, intelligence2025pi05, gr00tn1_2025, trilbmteam2025careful, geminiroboticsteam2025geminirobotics}, whose development depends on rapid iteration and large-scale benchmarking, this reliance has become a significant bottleneck.

Simulation offers a scalable and systematic alternative and is widely used for data generation and training~\cite{nvidia2024isaac, mujoco2012, Xiang_2020_SAPIEN, li2024behavior1k, Genesis, tedrake2019drake}. Yet it is far less common as a tool for policy evaluation, primarily due to poor sim-to-real correlation: a policy that performs well in simulation often fails to translate to similar real-world success. Narrowing this gap would allow simulation to serve as a trustworthy proxy for real-world testing, greatly accelerating development cycles. This raises the central question: \textit{how can we design simulators that are sufficiently realistic to evaluate robot policies with confidence?} To answer this question, we propose a framework for building high-fidelity simulators and investigate whether they can predict real-world policy performance reliably.

We identify two key factors for aligning simulation with reality: \textit{appearance} and \textit{dynamics}. On the appearance side, rendered scenes must closely match real-world observations. This is particularly challenging for policies that rely on wrist-mounted cameras, where simple green-screen compositing~\cite{li2024evaluating} is insufficient. We address this by leveraging 3D Gaussian Splatting (3DGS)~\cite{kerbl3Dgaussians}, which reconstructs photorealistic scenes from a single scan and supports rendering from arbitrary viewpoints. Beyond prior uses of 3DGS for simulation~\cite{abouchakra2025realissim, qureshi2024splatsim, li2024robogsim, barcellona2025dream}, we enhance it with automatic position and color alignment and object deformation handling, which are essential for closing the appearance gap.

Dynamics present another major source of sim-to-real discrepancy. Traditional simulators rely on low-dimensional parameter tuning, which is insufficient for deformable objects with many degrees of freedom. To address this challenge, we adopt PhysTwin~\cite{jiang2025phystwin}, a framework that reconstructs deformable objects as dense spring-mass systems optimized directly from object interaction videos. This approach yields efficient system identification while closely matching real-world dynamics.

We integrate these appearance and dynamics components into a unified simulator and expose it through a Gym-style interface~\cite{brockman2016openaigym}. We evaluate this framework on representative rigid- and soft-body manipulation tasks, including plush toy packing, rope routing, and T-block pushing, using widely adopted imitation learning algorithms: ACT~\cite{zhao2023learning}, Diffusion Policy (DP)~\cite{chi2023diffusionpolicy}, SmolVLA~\cite{shukor2025smolvla}, and Pi-0~\cite{black2024pi0}. By comparing simulated and real-world success rates and performing ablation studies, we observe a strong correlation and confirm that rendering and dynamics fidelity are both crucial to the trustworthiness of simulation-based evaluation.

In summary, our main contributions are:
(1)~A complete framework for evaluating robot policies in a Gaussian Splatting-based simulator using soft-body digital twins.
(2)~Empirical evidence that simulated rollouts strongly correlate with real-world success rates across representative tasks, using policies trained \textit{exclusively} on real-world data (no co-training).
(3)~A detailed analysis of design choices that improve the reliability of simulation as a predictor of real-world performance, offering guidance for future simulation-based evaluation pipelines.

%% file: sections/02-related-works.tex
\section{Related Works}

\subsection{Robot Policy Evaluation}
Evaluating robot policies is essential for understanding and comparing policy behaviors. Most systems are still evaluated directly in the real world~\cite{trilbmteam2025careful, octo_2023, pi0experimentwild, padalkar2023open, khazatsky2024droid, zhou2025autoeval, atreya2025roboarena}, but such evaluations are costly, time-consuming, and usually tailored to specific tasks, embodiments, and sensor setups. To enable more systematic study, prior works have introduced benchmarks, either in the real world through standardized hardware setups~\cite{Calli_2015, van2016robotic, 7583659, zhou2023trainofflinetestonline, dasari2022rb2roboticmanipulationbenchmarking} or in simulation through curated assets and task suites~\cite{taomaniskill3, james2019rlbenchrobotlearningbenchmark, li2024behavior1k, pmlr-v164-srivastava22a, puig2023habitat30cohabitathumans, robocasa2024, zhu2025robosuitemodularsimulationframework, mandlekar2023mimicgendatagenerationscalable, yang2025robotpolicyevaluationsimtoreal, wang2025roboevalroboticmanipulationmeets, badithela2025reliable}. Real-world benchmarks offer high fidelity but lack flexibility and scalability, while simulators often suffer from unrealistic dynamics and rendering, which limits their reliability as proxies for physical experiments. This is widely referred to as the ``sim-to-real gap''~\cite{peng2018sim, chebotar2019closing, openai2019solvingrubikscuberobot, ho2021retinaganobjectawareapproachsimtoreal}. We aim to narrow this gap by building a realistic simulator that combines high-quality rendering with faithful soft-body dynamics. Compared to SIMPLER~\cite{li2024evaluating} and RobotArena $\infty$~\cite{jangir2025robotarena} which relies on green-screen compositing, Ctrl-World~\cite{guo2025ctrl} which relies on 2D video world models, and Real-is-sim~\cite{abouchakra2025realissim} which focuses on rigid-body simulation, our method integrates Gaussian Splatting-based rendering with soft-body digital twins derived from interaction videos, eliminating the dependence on static cameras and providing more realistic appearance and dynamics.

\subsection{Physical Digital Twins}
Digital twins seek to realistically reconstruct and simulate real-world objects. Many existing frameworks rely on pre-specified physical parameters~\cite{liu2024physgen, chen2025physgen3d, jiang2024vr, xie2024physgaussian, qiu2024featuresplatting}, which limits their ability to capture complex real-world dynamics or leverage data from human interaction. While rigid-body twins are well studied~\cite{bianchini2025vysics, yang2025twintrackbridgingvisioncontact, abou-chakra2024physically, yu2016more}, full-order parameter identification for deformable objects remains challenging. Learning-based approaches have been proposed to capture such dynamics~\cite{pfaff2021learningmeshbasedsimulationgraph, zhang2024adaptigraph, zhang2025particle, tian2025diffusion}, but they often sacrifice physical consistency, which is critical for evaluating manipulation policies in contact-rich settings. Physics-based methods that optimize physical parameters from video observations~\cite{li2023pac, zhang2024physdreamer, zhong2024reconstruction, chen2025vid2sim} offer a more promising path. Among them, PhysTwin~\cite{jiang2025phystwin} reconstructs deformable objects as dense spring-mass systems directly from human-object interaction videos, achieving state-of-the-art realism and efficiency. Our work builds on PhysTwin and integrates its reconstructions with a Gaussian Splatting simulator to bridge the dynamics gap in policy evaluation.

\subsection{Gaussian Splatting Simulators}
Building simulators that closely match the real world requires high-quality rendering and accurate physics. Gaussian Splatting (3DGS)~\cite{kerbl3Dgaussians} has recently emerged as a powerful approach for scene reconstruction, enabling photorealistic, real-time rendering from arbitrary viewpoints~\cite{jiang2024vr, abou-chakra2024physically}. Several studies have demonstrated its potential in robotics, showing that 3DGS-based rendering can improve sim-to-real transfer for vision-based policies~\cite{qureshi2024splatsim, han2025re, escontrela2025gaussgym}, augment training datasets~\cite{li2024robogsim, barcellona2025dream, yu2025real2render2realscalingrobotdata, yang2025novel}, and enable real-to-sim evaluation~\cite{abouchakra2025realissim, jiang2025gsworld}. We extend this line of work by supporting soft-body interactions, incorporating PhysTwin~\cite{jiang2025phystwin} for realistic dynamics, and introducing automated position and color alignment, resulting in a complete and evaluation-ready simulator.

%% file: sections/03-method.tex
\section{Method}

\begin{figure*}[t]
    \centering
    \includegraphics[width=\linewidth]{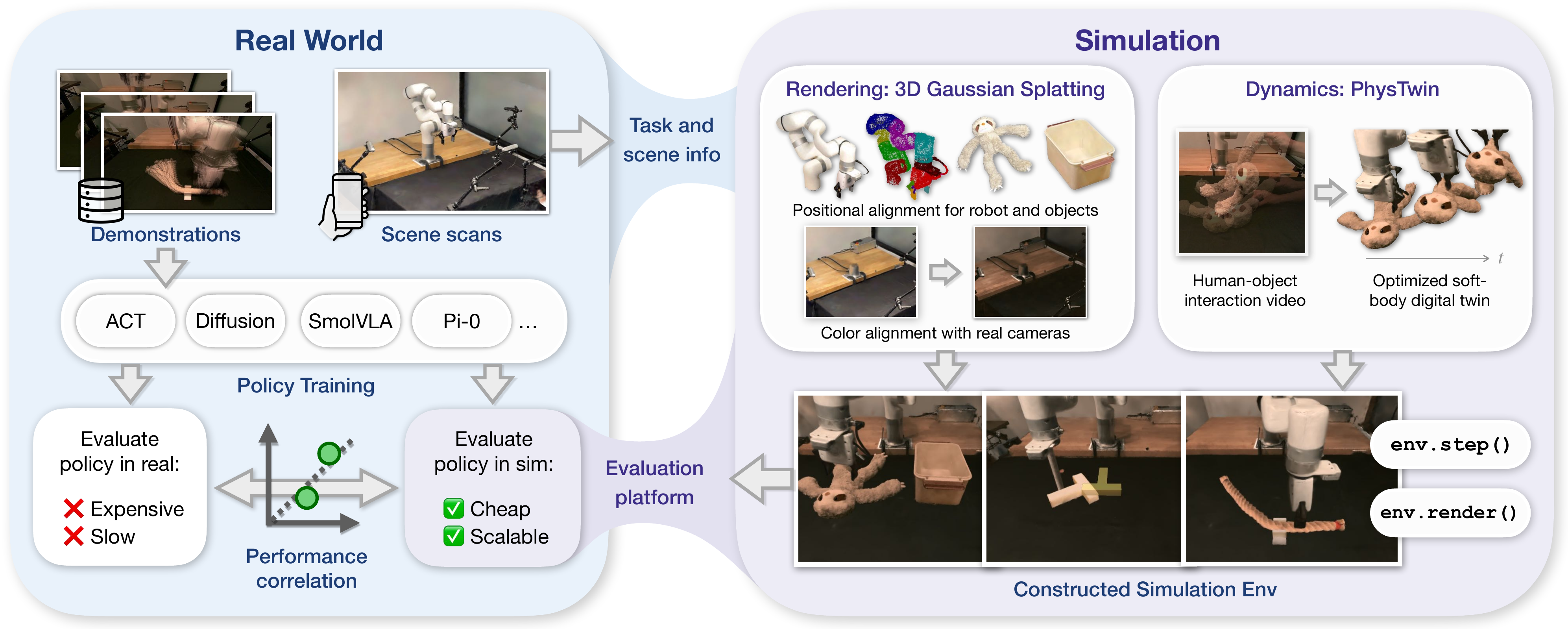}
    \vspace{-15pt}
    \caption{\small
    \textbf{Proposed framework for real-to-sim policy evaluation.} We present a pipeline that evaluates real-world robot policies in simulation using Gaussian Splatting-based rendering and soft-body digital twins. Policies are first trained on demonstrations collected by the real robot, and a phone scan of the workspace is used to reconstruct the scene via Gaussian Splatting. The reconstruction is segmented into robot, objects, and background, then aligned in position and color to enable photorealistic rendering. For dynamics, we optimize soft-body digital twins from object interaction videos to accurately reproduce real-world behavior. The resulting simulation is exposed through a Gym-style API~\cite{brockman2016openaigym}, allowing trained policies to be evaluated efficiently. Compared with real-world trials, this simulator is cheaper, reproducible, and scalable, while maintaining strong correlation with real-world performance.
    }
    \label{fig:method}
    \vspace{-10pt}
\end{figure*}

\subsection{Problem Definition}
We study the policy evaluation problem: \textit{Can a simulator reliably predict the real-world performance of visuomotor policies trained with real data?}
In a typical evaluation pipeline~\cite{kress2024robot,trilbmteam2025careful}, multiple policies are executed across controlled initial configurations in both simulation and the real world, and performance is measured through rollout-based metrics, typically expressed as scalar scores $u \in [0,1]$. The objective is to establish a strong correlation between simulated and real-world outcomes, represented by the paired set $\{(u_{i,\mathrm{sim}}, u_{i,\mathrm{real}})\}_{i=1}^N$, where $u_{i,\mathrm{sim}}$ and $u_{i,\mathrm{real}}$ denote the performance of the $i$-th policy in simulation and reality, respectively, and $N$ is the number of evaluated policies.

To achieve better performance correlation, one promising way is to build a simulator that yields consistent results with the real world. Formally, let $\{(s_t, o_t, a_t)\}_{t=1}^T$ denote the sequence of environment states $s_t$, robot observations $o_t$, and robot actions $a_t$ over a time horizon $T$. A simulator for policy evaluation should contain two core components:
(1)~\textit{Dynamics model:} $s_{t+1} = f(s_t, a_t)$, which predicts future states given the current state and robot actions.
(2)~\textit{Appearance model:} $o_t = g(s_t)$, 
which renders observations in the input modality required by the policy (e.g., RGB images).
Accordingly, the fidelity of simulation can be assessed along two axes: (i)~the accuracy of simulated dynamics, and (ii)~the realism of rendered observations.

In this work, we address both axes by jointly reducing the visual gap and the dynamics gap. We employ physics-informed reconstruction of soft-body digital twins to align simulated dynamics with real-world object behavior, and use high-resolution Gaussian Splatting as the rendering engine to generate photorealistic observations. The following sections describe these components in detail, and an overview of the full framework is shown in Figure~\ref{fig:method}.

\subsection{Preliminary: PhysTwin}
We adopt the PhysTwin~\cite{jiang2025phystwin} digital twin framework, which reconstructs and simulates deformable and rigid objects from video using a dense spring-mass system. Each object is represented as a set of mass nodes connected by springs, with springs formed between each pair of nodes within a distance threshold $d$. The node positions evolve according to Newtonian dynamics.

To capture the behavior of diverse real-world deformable objects with varying stiffness, friction, and other material properties, PhysTwin employs a real-to-sim pipeline that jointly optimizes a set of physical parameters, including the spring threshold $d$ and per-spring stiffness coefficients $Y$. The optimization is performed from a single video of a human interacting with the object by hand: human hand keypoints are tracked and attached to the spring-mass system as kinematic control points, and system parameters are tuned to minimize the discrepancy between tracked object motions in the video and their simulated counterparts. For rigid bodies, $Y$ is fixed to a large value to suppress deformation. We adopt this same real-to-sim process for system identification of the objects that interact with the robot (plush toy, rope, and T-block).

\subsection{Real-to-Sim Gaussian Splatting Simulation}
We now describe the construction of our Gaussian Splatting-based simulator. Our approach addresses two complementary goals: (i)~closing the visual gap through GS scene reconstruction, positional alignment, and color alignment, and (ii)~closing the dynamics gap through physics-based modeling and deformation handling. 

\subsubsection{GS Construction}
We begin by acquiring the appearance of each object of interest using Scaniverse~\cite{scaniverse}, an iPhone app that automatically generates GS reconstructions from video recordings. In a tabletop manipulation scene, we first scan the static robot workspace, including the robot, table, and background, then scan each experimental object individually. The resulting reconstructions are segmented into robot, objects, and background using the SuperSplat~\cite{supersplat2025} interactive visualizer. This reconstruction step is required only once per task. 

\subsubsection{Positional Alignment}
After obtaining GS reconstructions of the static background, robot, PhysTwin object, and other static objects, we align all components to the reference frames: the robot base frame and canonical object frames. PhysTwin objects and static meshes are aligned to their corresponding PhysTwin particle sets and object 3D models by applying a relative 6-DoF transformation. For the robot, we automatically compute the transformation between the reconstructed GS model and ground truth robot points (generated from its URDF) using a combination of Iterative Closest Point (ICP)~\cite{4767965} and RANSAC~\cite{10.1145/358669.358692}. We use 2,000 points per link to ensure sufficient coverage of link geometry. Because the background GS is in the same frame as the robot GS, we apply the same transformation estimated by ICP.

To enable the simulation of the static robot GS, we associate each Gaussian kernel with its corresponding robot link through a link segmentation process. After ICP alignment, each kernel is assigned to a link by finding its nearest neighbor in the sampled robot point cloud and inheriting that point’s link index. This process is applied to all links, including the gripper links, allowing us to render continuous arm motion as well as gripper opening and closing. The same procedure generalizes naturally to other robot embodiments with available URDF models.

\subsubsection{Color Alignment}
A major contributor to the visual gap in GS renderings is that reconstructed scenes often lie in a different color space from the policy’s training data, leading to mismatched pixel color distributions, which can affect policy performance. In our setting, GS reconstructions inherit the color characteristics of iPhone video captures, while policies are trained in the color space of the robot’s cameras (e.g., Intel RealSense, which is known to introduce color shifts). To close this gap, we design a color transformation that aligns GS colors to the real camera domain.

We perform this alignment directly in RGB space. First, we render images from the scene GS at the viewpoints of the fixed real cameras, using the original Gaussian kernel colors and opacities. Next, we capture real images from the same viewpoints, forming paired data for optimization. We then solve for a transformation function $f$ that minimizes the pixel-wise color discrepancy:
\begin{equation}
    f^* = \arg\min_{f\in \mathcal{F}} \frac{1}{N}\sum_{i=1}^N \|f(p_i) - q_i\|_2, \ \ p_i \in I_{GS}, \ \ q_i\in I_{RS},
\end{equation}
where $I_{GS}$ and $I_{RS}$ denote GS renderings and real camera captures, $N$ is the number of pixels, $p_i$ and $q_i$ are corresponding RGB values, and $\mathcal{F}$ is the function space. We parameterize $\mathcal{F}$ as the set of degree-$d$ polynomial transformations:
\begin{gather}
    f = \{f_i\}_{i=1}^d, \ \ f_i \in\mathbb{R}^3,
    \\
    f(p_i) = [f_0\ f_1\ \cdots\ f_{d}] \cdot [1\ p_i\ \cdots\ p_i^{d}]^T,
\end{gather}
which reduces the problem to a standard least-squares regression. We solve it using Iteratively Reweighted Least Squares (IRLS)~\cite{green1984iteratively} to improve robustness to outliers. Empirically, we find that a quadratic transform ($d=2$) offers the best trade-off between expressivity and overfitting.

\subsubsection{Physics and Deformation}
With GS reconstruction and alignment mitigating the rendering gap, the physics model must accurately capture real-world dynamics. We use a custom physics engine built on NVIDIA Warp~\cite{warp2022}, extending the PhysTwin~\cite{jiang2025phystwin} spring-mass simulator to support collisions with both robot end-effectors and objects in the environment.
For grasping soft-body digital twins, we avoid the common but unrealistic practice of fixing object nodes to the gripper. Instead, we model contact purely through frictional interactions between gripper fingers and the object. The gripper closing motion halts automatically once a specified total collision-force threshold is reached, yielding more realistic and stable grasps.

At each simulation step, the updated robot and environment states from the physics engine are propagated to the Gaussian kernels. For rigid bodies, including objects and robot links, kernel positions and orientations are updated using the corresponding rigid-body transformations. For deformable objects, following PhysTwin~\cite{jiang2025phystwin}, we apply Linear Blend Skinning (LBS)~\cite{10.1145/1276377.1276478} to transform each kernel based on the underlying soft-body deformation. 

Overall, with GS rendering, the physics solver, and LBS-based deformation being the major computational steps, our simulator runs at 5 to 30 FPS on a single GPU, depending on the robot-object contact states. By eliminating the overhead of real-world environment resets and leveraging multi-GPU parallelization, we empirically achieve evaluation speeds several times faster than real-world execution.

\subsection{Policy Evaluation}
To evaluate visuomotor policies in our simulator, we first design tasks and perform real-world data collection and policy training. Demonstrations are collected through human teleoperation using GELLO~\cite{wu2023gello}, after which we scan the scene to construct the corresponding simulation environments. All policies are trained \textit{exclusively} on real data (i.e., no co-training between simulation and reality). To improve consistency and reduce variance, we follow the practice of Kress-Gazit et al.~\cite{kress2024robot} by defining a fixed set of initial object configurations for each task and performing evaluations in both simulation and the real world. In the real world, we use a real-time visualization tool that overlays simulated initial states onto live camera streams, enabling operators to accurately and consistently reproduce the starting configurations.

Policy performance $u$ is measured in terms of binary task success rates: in the real world, success is determined by human evaluators, while in simulation, task-specific criteria are automatically computed from privileged simulation states. In this work, we evaluate the performance of several state-of-the-art imitation learning algorithms, as well as checkpoints from different training stages for each network. Notably, the simulator is readily extensible to other policy types, as we package the entire system into the widely adopted Gym environment API~\cite{brockman2016openaigym}. We are committed to open-sourcing our implementation to encourage community adoption and enable scalable, reproducible policy evaluation.

%% file: sections/04-experiments.tex
\section{Experiments}

\begin{figure*}[t]
    \centering
    \includegraphics[width=\linewidth]{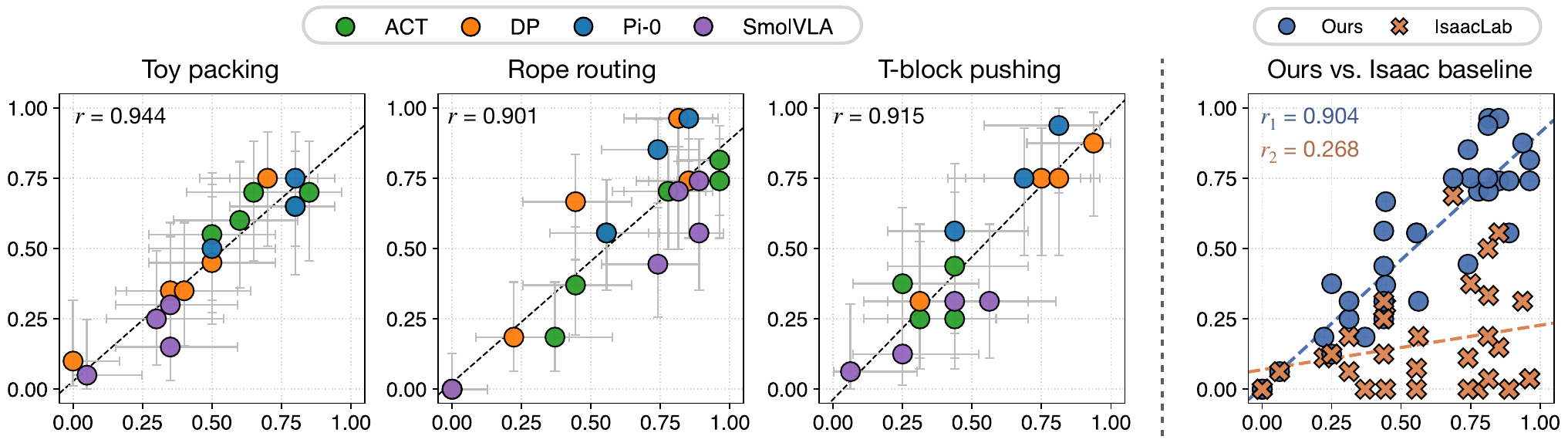}
    \vspace{-15pt}
    \caption{\small \textbf{Correlation between simulation and real-world policy performance.} 
    \textit{Left:} Simulation success rates ($y$-axis) vs. real-world success rates ($x$-axis) for toy packing, rope routing, and T-block pushing, across multiple state-of-the-art imitation learning policies and checkpoints. The tight clustering along the diagonal indicates that, even with binary success metrics, our simulator faithfully reproduces real-world behaviors across tasks and policy robustness levels.
    \textit{Right:} Compared with IsaacLab, which models rope routing and push-T tasks, our approach yields substantially stronger sim-to-real correlation, highlighting the benefit of realistic rendering and dynamics.}
    \label{fig:corr}
    \vspace{-15pt}
\end{figure*}

\begin{figure}[t]
    \centering
    \includegraphics[width=\linewidth]{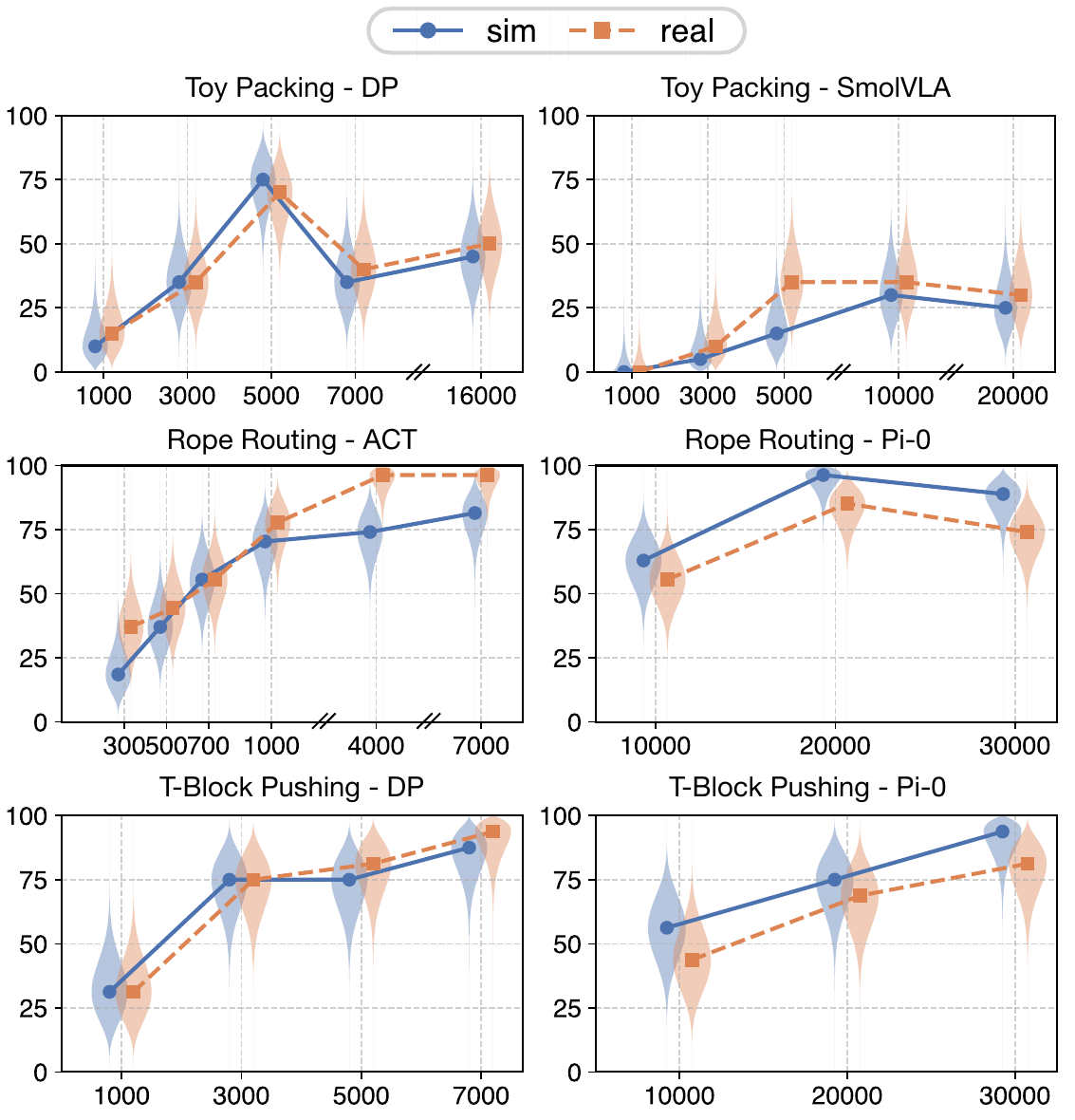}
    \vspace{-15pt}
    \caption{\textbf{Per-policy, per-task performance across training.} $x$-axis: training iterations, $y$-axis: success rates. Simulation (blue) and real-world (orange) success rates are shown across iterations. Unlike Figure~\ref{fig:corr}, which aggregates across policies, this figure shows unrolled curves for each task-policy pair. Improvements in simulation consistently correspond to improvements in the real world, establishing a positive correlation and demonstrating that our simulator can be a reliable tool for evaluating/selecting policies.}
    \label{fig:curve}
    \vspace{-10pt}
\end{figure}

In this section, we test the performance of imitation learning policies in both the real world and our simulation environment to examine the correlation. We aim to address the following questions:
(1)~How strongly do the simulation and real-world performance correlate?
(2)~How critical are rendering and dynamics fidelity for improving this correlation?
(3)~What practical benefits can the correlation provide?

\subsection{Experiment Setup}
\subsubsection{Tasks}
We evaluate policies on three representative manipulation tasks involving both deformable and rigid objects:

\begin{itemize}
\item \textit{Toy packing}: The robot picks up a plush sloth toy from the table and packs it into a small plastic box. A trial is considered successful only if the toy’s arms, legs, and body are fully contained within the box, with no parts protruding.
\item \textit{Rope routing}: The robot grasps a cotton rope, lifts it, and routes it through a 3D-printed clip. Success is defined by the rope being fully threaded into the clip.
\item \textit{T-block pushing (push-T)}: A 3D-printed T-shaped block is placed on the table. Using a vertical cylindrical pusher, the robot must contact the block and then translate and reorient it to match a specified target pose.
\end{itemize}

Both the toy packing and rope routing tasks are challenging because the small tolerances of the box and clip require the policy to leverage visual feedback. Similarly, in push-T, the policy must infer the block’s pose from images to achieve the required translation and reorientation.

\subsubsection{Evaluation}
To reduce variance and ensure systematic evaluation, we initialize scenes from a fixed set of configurations shared between the simulation and the real world. These initial configurations are generated in our simulator by constructing a grid over the planar position $(x, y)$ and rotation angle $\theta$ of objects placed on the table. The grid ranges are chosen to ensure that the evaluation set provides coverage comparable to the training distribution. In the real world, objects are positioned to replicate the corresponding grid states. We use an evaluation set size of 20, 27, and 16 for toy packing, rope routing, and push-T, respectively.

We use binary success criteria for all tasks. Following~\cite{li2024evaluating}, we quantify the alignment between simulation and real-world performance using the Mean Maximum Rank Variation (MMRV) and the Pearson correlation coefficient ($r$).

The number of evaluation episodes plays a critical role in the uncertainty of measured success rates~\cite{trilbmteam2025careful}. To capture this variability, we report uncertainty in our results using the Clopper–Pearson confidence interval (CI). We also visualize the Bayesian posterior of policy success rates under a uniform Beta prior with violin plots.

We evaluate four state-of-the-art imitation learning policies: ACT~\cite{zhao2023learning}, DP~\cite{chi2023diffusionpolicy}, Pi-0~\cite{black2024pi0}, and SmolVLA~\cite{shukor2025smolvla}. The real-world setup consists of a single UFactory xArm 7 robot arm equipped with two calibrated Intel RealSense RGB-D cameras: a D405 mounted on the robot wrist and a D455 mounted on the table as a fixed external camera. All policies take as input images from both camera views, along with the current end-effector state. For push-T, the end-effector state includes only the 2D position $(x, y)$; for the other tasks, it additionally includes the position, rotation, and gripper openness. Across all tasks, we collect 39-60 successful demonstrations via teleoperation using GELLO~\cite{wu2023gello}. Training is performed using the open-source LeRobot~\cite{cadene2024lerobot} implementation, except for Pi-0, where we adopt the original implementation~\cite{black2024pi0} for better performance.

\subsection{Baseline}
As a baseline, we use NVIDIA IsaacLab~\cite{nvidia2024isaac} as the simulation environment. Robot and environment assets are imported and aligned in position and color to match the real-world setup. IsaacLab provides a general-purpose robot simulation framework built on the PhysX physics engine, but its support for deformable objects remains limited. For ropes, we approximate deformable behavior using an articulated chain structure. However, for the plush toy, realistic grasping and deformation could not be stably simulated, making task completion infeasible; we therefore excluded this task from our quantitative comparisons.

\subsection{Sim-and-Real Correlation}
Figure~\ref{fig:corr} (left) shows the performance of all policy checkpoints in both simulation and the real world. We observe a strong correlation: policies that achieve higher success rates in reality also achieve higher success rates in our simulator, consistently across architectures and tasks. Figure~\ref{fig:corr} (right) further highlights that our simulator achieves stronger correlation than the IsaacLab baseline~\cite{nvidia2024isaac}. This is also confirmed by the quantitative results in Table~\ref{tab:corr}, with our simulator achieving a Pearson coefficient $r > 0.9$ for all policies. By contrast, the baseline yields only $r = 0.649$ on push-T, and an even lower $r = 0.237$ on rope routing as a result of the larger dynamics gap. The low MMRV value for the IsaacLab rope routing task arises from its consistently low success rates, which in turn produce fewer ranking violations.

\begin{figure*}[t]
    \centering
    \includegraphics[width=\linewidth]{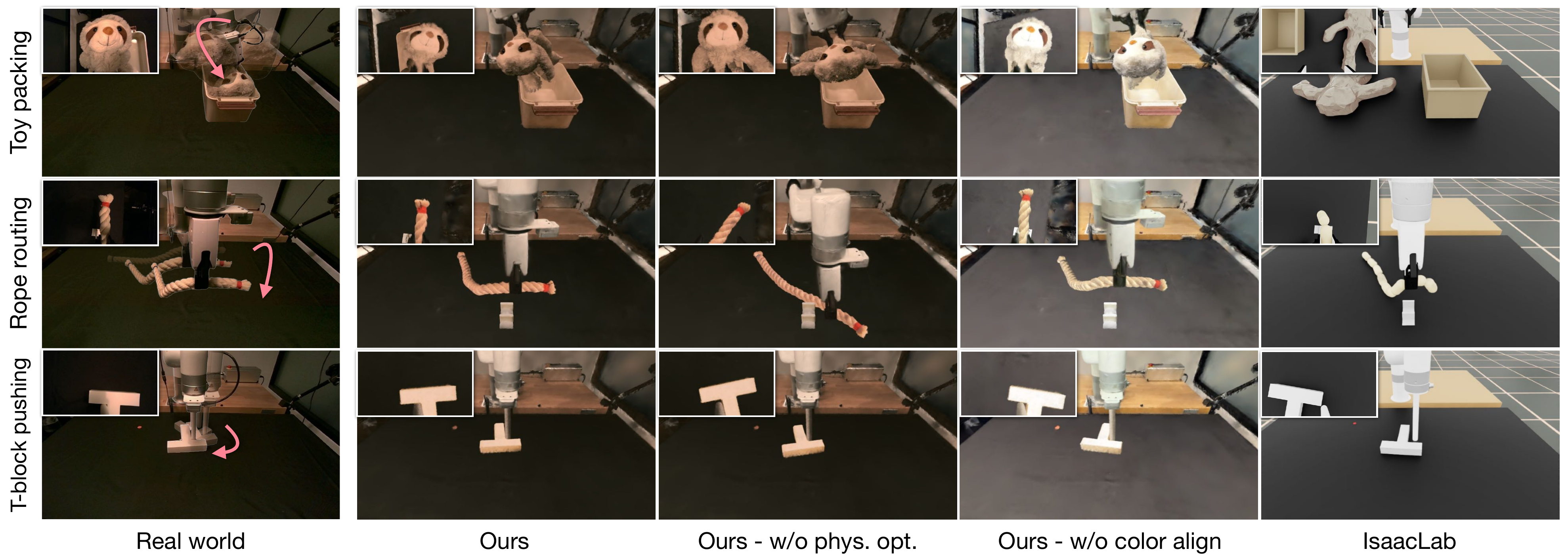}
    \vspace{-15pt}
    \caption{\small
    \textbf{Comparison of rendering and dynamics quality.} Real-world observations (left) compared with our method, two ablations, and the IsaacLab baseline across three tasks. From right to left, visual and physical fidelity progressively improve. Without physics optimization, object dynamics deviate, causing failures such as the toy’s limbs not fitting into the box or the rope slipping before routing. Without color alignment, rendered images exhibit noticeable color mismatches. The IsaacLab baseline (rightmost) shows lower realism in both rendering and dynamics compared to our approach.}
    \label{fig:vis}
    \vspace{-15pt}
\end{figure*}

\subsection{Policy Performance Analysis}
Figure~\ref{fig:curve} further illustrates per-policy, per-task performance curves across training iterations. We observe that simulation success rates generally follow the same progression as real-world success rates, further highlighting the correlation. For example, in the toy packing-DP case, both simulation and real success rates peak at iteration 5,000 and decline significantly by iteration 7,000. Similarly, in the rope routing-Pi-0 case, performance peaks around iteration 20,000. These results suggest that our simulator can be used as a practical tool for monitoring policy learning dynamics, selecting checkpoints for real-world testing, and setting approximate expectations for real-world performance.

In cases where simulation and real success rates do not overlap, such as toy packing-SmolVLA and rope routing-ACT, the simulator still captures the correct performance trend, even if the absolute success rates differ. We attribute these discrepancies to residual gaps in visual appearance and dynamics, as well as variance from the limited number of evaluation episodes (16–27 per checkpoint).

\begin{table}[t]
\scriptsize
\begin{tabular}{@{}lcccccc@{}}
\toprule
                       & \multicolumn{2}{c}{Toy packing}   & \multicolumn{2}{c}{Rope routing}  & \multicolumn{2}{c}{T-block pushing} \\ \midrule
                       & MMRV$\downarrow$ & $r\uparrow$    & MMRV$\downarrow$ & $r\uparrow$    & MMRV$\downarrow$  & $r\uparrow$     \\ \midrule
IsaacLab~\cite{nvidia2024isaac}  & -          & -        & 0.270            & 0.237          & 0.196             & 0.649           \\
Ours w/o color & 0.200            & 0.805          & 0.343            & 0.714          & 0.354             & 0.529           \\
Ours w/o phys.  & 0.241            & 0.694          & 0.248            & 0.832          & \textbf{0.100}                  & 0.905                \\
Ours                   & \textbf{0.076}   & \textbf{0.944} & \textbf{0.174}   & \textbf{0.901} & 0.108    & \textbf{0.915}  \\ \bottomrule
\end{tabular}
\caption{\small
\textbf{Quantitative comparison of correlation.} \textit{Ours w/o color}: our method without color alignment. \textit{Ours w/o phys.}: our method without physics optimization. Lower MMRV indicates fewer errors in ranking policy performance, while higher $r$ reflects stronger statistical correlation. Best results are highlighted in bold. 
}
\label{tab:corr}
\vspace{-15pt}
\end{table}

\subsection{Ablation Study}
To measure the importance of the rendering and dynamics realism for our Gaussian Splatting simulator, we perform ablation studies on the correlation metrics MMRV and $r$. We provide two ablated variants of our simulation:
\begin{itemize}
    \item \textit{Ours w/o color alignment}: we skip the color alignment step in simulation construction and use the original GS colors in the iPhone camera space, creating a mismatch in the appearance.
    \item \textit{Ours w/o physics optimization}: instead of using the fully-optimized spring stiffness $Y$, we use a global stiffness value shared across all springs. The global value is given by the gradient-free optimization stage in PhysTwin~\cite{jiang2025phystwin}. For push-T, we keep its rigidity and change its friction coefficients with the ground and the robot to create a mismatch in dynamics.
\end{itemize}

Figure~\ref{fig:vis} presents a visual comparison between our simulator, its ablated variants, and the baseline, using the same policy model and identical initial states. Our full method achieves the best rendering and dynamics fidelity, resulting in policy rollouts that closely match real-world outcomes. In contrast, the w/o physics optimization variant produces inaccurate object dynamics, while the w/o color alignment variant shows clear color mismatches.

Empirically, both dynamics and appearance mismatches lead to deviations between simulated and real policy rollouts, though policies exhibit different sensitivities to each type of gap. For example, in the rope routing task, the rope fails to enter the clip when stiffness is mis-specified (\textit{w/o physics optimization}). In the push-T task, color discrepancies alter the robot’s perception, causing it to push the block differently (\textit{w/o color alignment}).

Table~\ref{tab:corr} details the quantitative results. Overall, our full method achieves the highest correlation values, outperforming the ablated variants. In particular, lower MMRV values reflect more accurate policy ranking, while higher Pearson correlation coefficients ($r$) indicate stronger and more consistent correlations without being influenced by outlier points. 

%% file: sections/05-conclusion.tex
\section{Conclusion}

In this work, we introduced a framework for evaluating robot manipulation policies in a simulator that combines Gaussian Splatting-based rendering with real-to-sim digital twins for deformable object dynamics. By addressing both appearance and dynamics, our simulator narrows the sim-to-real gap through physics-informed reconstruction, positional and color alignment, and deformation-aware rendering. 

We demonstrated the framework on representative deformable and rigid body manipulation tasks, evaluating several state-of-the-art imitation learning policies. Our experiments show that policy success rates in simulation exhibit strong correlations with real-world outcomes ($r > 0.9$). Further analysis across highlights that our simulator can predict policy performance trends, enabling it to serve as a practical proxy for checkpoint selection and performance estimation. We found that both physics optimization and color alignment are critical for closing policy performance gaps. 

In future work, scaling both simulation and evaluation to larger task and policy sets could provide deeper insights into the key design considerations for policy evaluation simulators. Moreover, our real-to-sim framework can be generalized to more diverse environments, supporting increasingly complex robot manipulation tasks.

%% file: sections/06-appendix.tex
\section{Additional Technical Details}
\label{appendix:details}

\subsection{Platform and Tasks}

\subsubsection{Robot Setup}
We use a UFactory xArm 7 robot mounted on a tabletop. The robot arm has 7 degrees of freedom. The robot end-effector can be interchanged between the standard xArm gripper and a custom 3D-printed pusher, depending on the task. Two Intel RealSense RGB-D cameras are connected to the robot workstation: a D455 fixed on the table overlooking the workspace, and a D405 mounted on the robot wrist via a custom 3D-printed clip. To ensure consistent appearance between real and simulated observations, we fix the white balance and exposure settings of both cameras.

\subsubsection{Data Collection}
We use GELLO for data collection. GELLO~\cite{wu2023gello} streams high-frequency joint-angle commands to the robot, which we execute using joint-velocity control for smooth motion tracking. At each timestep, the robot computes the difference between the commanded and measured joint angles, then sets each joint’s target angular velocity proportional to this delta. To prevent abrupt movements, the velocity vector is normalized such that its total $\ell_2$ norm does not exceed a predefined limit. This approach enables stable and continuous trajectory following without jerky motions. During policy evaluation, we apply the same control strategy, ensuring that the policy outputs are tracked consistently in both real and simulated environments.

\begin{figure}[t]
    \centering
    \includegraphics[width=\linewidth]{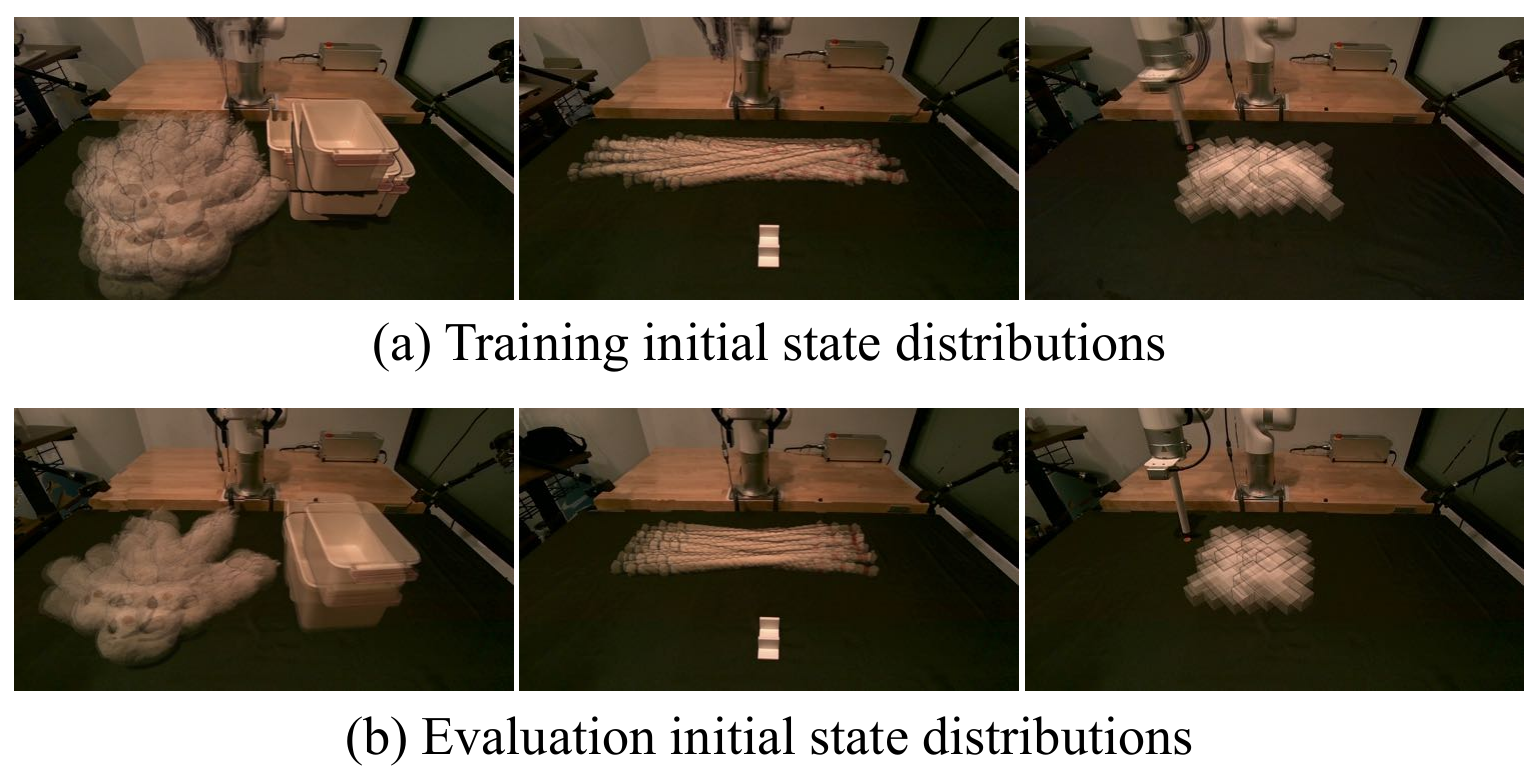}
    \vspace{-8pt}
    \caption{\textbf{Training and evaluation data distributions.}  
    Top: spatial coverage of initial states in the training set. Bottom: the corresponding coverage in the evaluation set.}
    \label{fig:coverage}
\end{figure}

\begin{table}[t]
\centering
\footnotesize
\setlength{\tabcolsep}{8pt}
\begin{tabular}{lcc}
\toprule
\textbf{Name} & \textbf{Dynamics Type} & \textbf{3D Representation} \\
\midrule
xArm-gripper-tabletop   & Articulated+Fixed & GS+URDF+Mesh \\
xArm-pusher-tabletop    & Articulated+Fixed & GS+URDF+Mesh \\
Plush sloth  & Deformable & GS+PhysTwin \\
Rope         & Deformable & GS+PhysTwin \\
T-block      & Rigid      & GS+PhysTwin \\
Box          & Fixed      & GS+Mesh     \\
Clip         & Fixed      & GS+Mesh     \\
\bottomrule
\end{tabular}
\caption{\textbf{Simulation assets.} 
Each row corresponds to an individual Gaussian Splatting scan, specifying its dynamics type in simulation and the 3D representation used for physical simulation and rendering. These assets are combined to instantiate all three manipulation tasks within the simulator.}
\label{tab:assets}
\vspace{-10pt}
\end{table}

\subsubsection{Task Definition}
To evaluate the effectiveness of our simulator, we select a set of rigid- and soft-body manipulation tasks that require the policy to leverage object dynamics while incorporating visual feedback. The formulation and setup of each task are described as follows.

\paragraph{Toy Packing} The robot grasps the plush toy by one of its limbs, lifts it above the box, and adjusts its pose such that the arm and leg on one side hang into the box. The robot then tilts the toy slightly to allow the other side’s limbs to enter, before lowering it further to pack the toy snugly inside the box. Because the box is intentionally compact, the robot must adapt to the toy’s pose to successfully execute the packing motion without leaving any limbs protruding over the box edges. A total of 39 human demonstration episodes are recorded for this task.

\paragraph{Rope Routing} The robot grasps one end of the rope (marked with red rubber bands), lifts it, and positions it above the cable holder before lowering it to gently place the rope into the slot. Because the rope–holder contact point is offset from the grasp location, the rope dynamics play a critical role in determining the appropriate displacement and trajectory required for successful placement. A total of 56 human demonstration episodes are collected for this task.

\paragraph{T-block Pushing} The robot begins with the pusher positioned above an orange marker on the table, while the end-effector’s $z$-coordinate remains fixed throughout the motion. The robot must move to the T-block’s location and push it toward a predefined goal region. The goal is not physically marked in the workspace but is visualized as a yellow translucent mask overlaid on the fixed-camera images. The initial positions and orientations of the T-block are randomized, and a total of 60 human demonstration episodes are collected for this task.

\begin{figure*}[t]
    \centering
    \includegraphics[width=\linewidth]{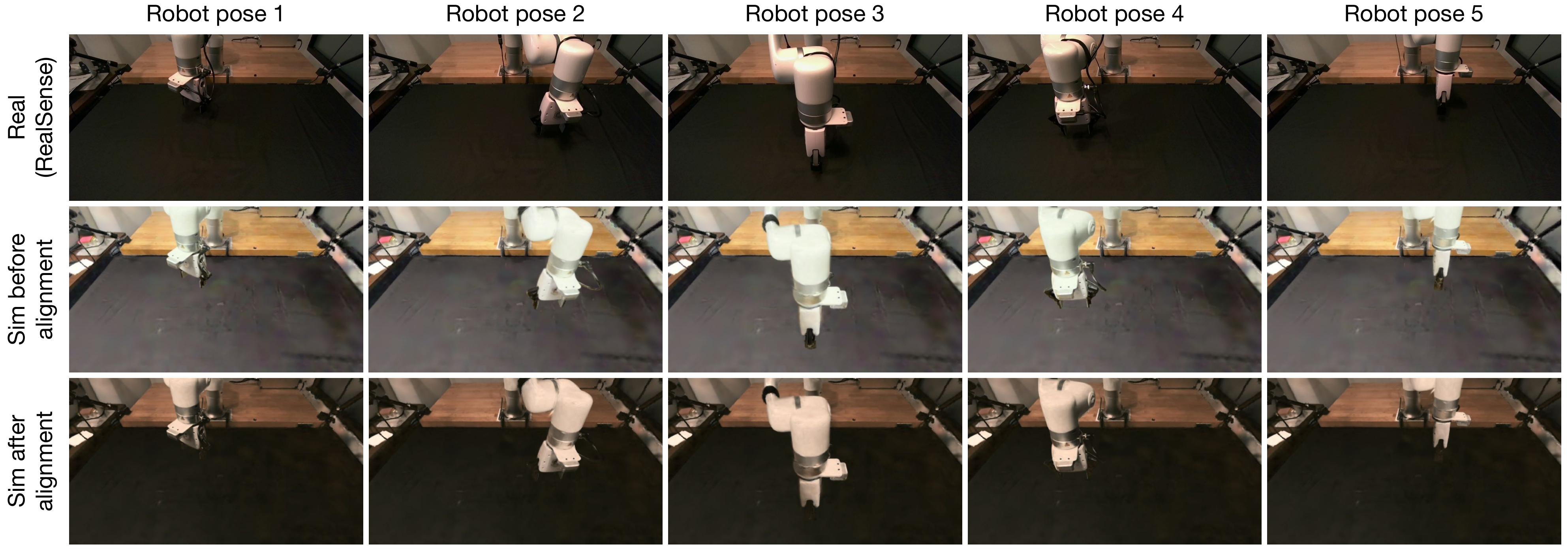}
    \vspace{-15pt}
    \caption{\small
    \textbf{Color alignment.} Five image pairs used for the color alignment process are shown. \textit{Top:} real images captured by the RealSense cameras. \textit{Middle:} raw Gaussian Splatting renderings with the robot posed identically to the real images. \textit{Bottom:} GS renderings after applying the optimized color transformation, showing improved consistency with real-world color appearance.
    }
    \label{fig:color_align}
    \vspace{-10pt}
\end{figure*}

\begin{figure}[t]
    \centering
    \includegraphics[width=\linewidth]{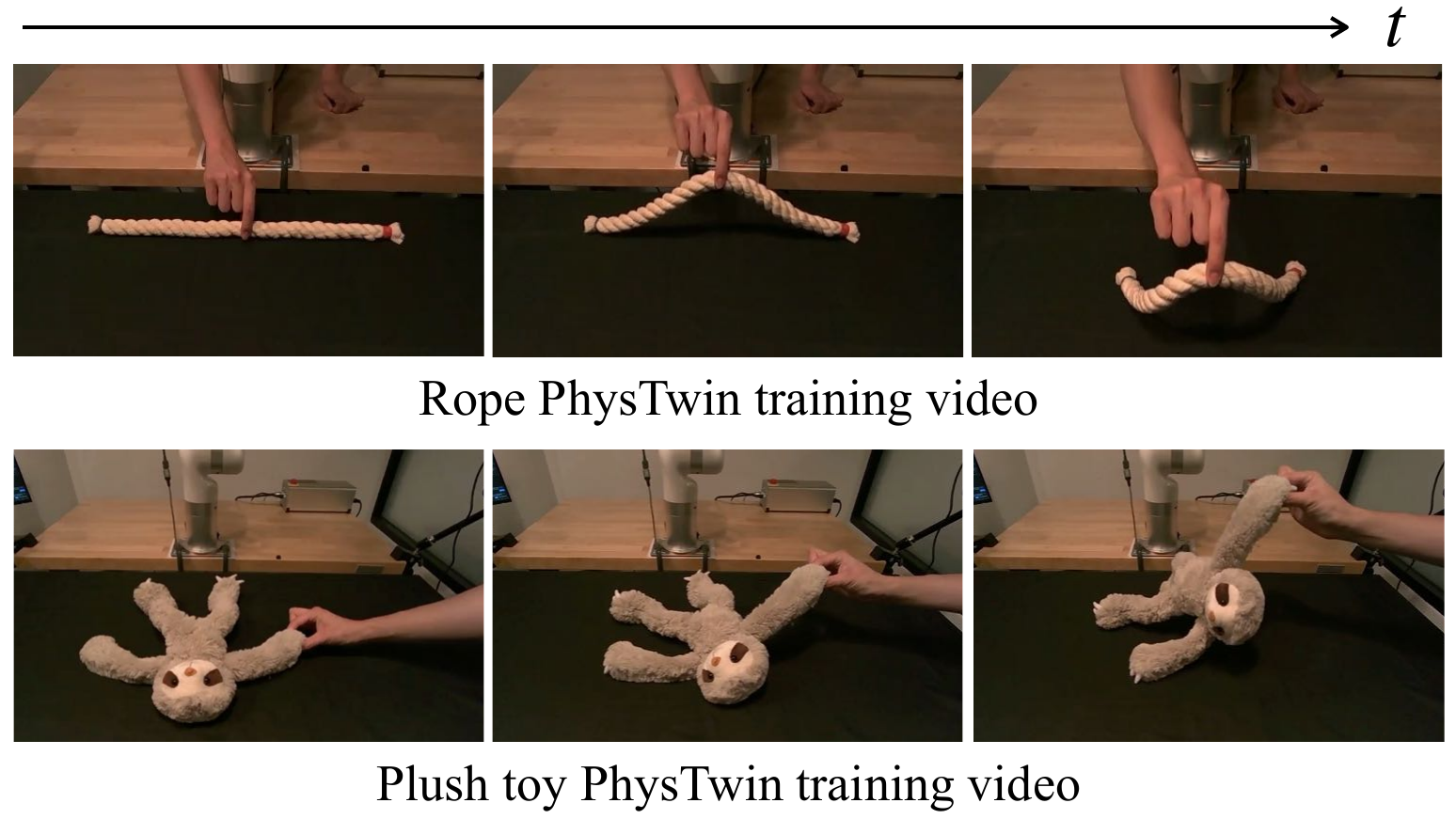}
    \vspace{-8pt}
    \caption{\textbf{PhysTwin training videos.} A few representative camera frames are shown for each training video, where a human subject interacts with the deformable object by hand. These videos are used by PhysTwin to reconstruct the object's geometry and estimate its physical parameters for building the digital twin models.}
    \label{fig:phystwin}
    \vspace{-10pt}
\end{figure}

\begin{algorithm2e}[t]
\caption{\textbf{Simulation Loop}}
\label{alg:sim}
\KwData{PhysTwin particle positions and velocities $x, v$, PhysTwin spring-mass parameters $P$, robot mesh $R$, robot motion $a$, static meshes $M_{1:k}$, ground plane $L$, total timestep $T$, substep count $N$, Gaussians $G$}
\For{$t \gets 0$ \KwTo $T-1$}{
    $x^*, v^* = x_{t}, v_{t}$\\
    $R^*_{1:N} = \text{interpolate\_robot\_states}(R_t, a_t)$\\
    \For{$\tau \gets 0$ \KwTo $N-1$}{
        $v^* = \text{step\_springs}(x^*, v^*, P)$\\
        $v^* = \text{self\_collision}(x^*, v^*, P)$\\
        $x^*, v^* = \text{robot\_mesh\_collision}(x^*, v^*, R_\tau, a_\tau)$\\
        \For{$i \gets 1$ \KwTo $k$}{
            $x^*, v^* = \text{fixed\_mesh\_collision}(x^*, v^*, M_i)$\\
        }
        $x^*, v^* = \text{ground\_collision}(x^*, v^*, L)$\\
    }
    $x_{t+1}, v_{t+1} = x^*, v^*$\\
    $R_{t+1} = R_{N}^*$\\
    $G_{t+1} = \text{renderer\_update}(G_{t}, x_t, x_{t+1}, R_{t}, R_{t+1})$
}
\end{algorithm2e}

\subsection{Simulation}

\subsubsection{Assets}
A summary of the simulation assets used in our experiments is provided in Table~\ref{tab:assets}. Each asset corresponds to a single Gaussian Splatting reconstruction followed by a pose alignment process.

\subsubsection{Positional Alignment}
To align the robot-scene Gaussian Splatting scan with the robot’s URDF model, we first perform a coarse manual alignment in SuperSplat~\cite{supersplat2025} to roughly match the origins and orientations of the $x$, $y$, and $z$ axes. Next, we manually define a bounding box to separate the robot Gaussians from the scene Gaussians. We then apply ICP registration between two point clouds: one formed by the centers of the robot Gaussians, and the other by uniformly sampled surface points from the robot URDF mesh. The resulting rigid transformation is applied to the entire GS, ensuring that both the robot and scene components are consistently aligned in the unified coordinate frame.

\subsubsection{Color Alignment}
The robot–scene scan has the most significant influence on the overall color profile of the rendered images. To align its appearance with the RealSense color space, we apply Robust IRLS with Tukey bi-weight to estimate the color transformation. We use five images of resolution 848$\times$480 for this optimization.
To mitigate the imbalance between the dark tabletop and the bright robot regions, each pixel is weighted by the norm of its RGB values, giving higher weight to high-brightness pixels in the least-squares loss. The optimization is run for 50 iterations. Figure~\ref{fig:color_align} visualizes the input images and the resulting color alignment.

\subsubsection{PhysTwin Training}
We use the original PhysTwin~\cite{jiang2025phystwin} codebase for training the rope and sloth digital twins. PhysTwin requires only a single multi-view RGB-D video to reconstruct object geometry and optimize physical parameters. For data capture, we record using three fixed Intel RealSense D455 cameras. The videos for the two objects are visualized in Figure~\ref{fig:phystwin}.
For the T-block pushing task, since it is a rigid object, we construct the PhysTwin object by uniformly sampling points within the mesh, connecting them with springs using a connection radius of 0.5 and a maximum of 50 neighbors, and assigning a uniform spring stiffness of $3\times10^4$ to all connections. This setup ensures that the object behaves like a rigid body.

\subsubsection{Simulation Loop}
The simulation loop, including robot action processing, PhysTwin simulation, collision handling, and renderer updates, is summarized in Algorithm~\ref{alg:sim}.

\begin{table}[t]
\centering
\footnotesize
\setlength{\tabcolsep}{8pt}
\begin{tabular}{lcccc}
\toprule
\textbf{Model} & \textbf{Visual} & \textbf{State} & \textbf{Action} & \textbf{Relative?}\\
\midrule
ACT   & mean–std & mean–std & mean–std & False \\
DP    & mean–std & min–max  & min–max  & False \\
SmolVLA & identity & mean–std & mean–std & True \\
Pi-0  & mean–std & mean–std & mean–std & True \\
\bottomrule
\end{tabular}
\caption{\textbf{Normalization schemes across models.} 
Columns indicate the normalization applied to each modality (visual, state, and action) and whether the model operates in a \textit{relative action space}. 
\textit{Mean–std} denotes standardization to zero mean and unit variance, while \textit{min–max} scales values to $[-1,1]$.}
\label{tab:normalization}
\end{table}

\begin{table}[t]
\centering
\footnotesize
\setlength{\tabcolsep}{6pt}
\begin{tabular}{ll|ll}
\toprule
\multicolumn{2}{c|}{\textbf{Color Transformations}} &
\multicolumn{2}{c}{\textbf{Spatial Transformations}} \\
\cmidrule(lr){1-2} \cmidrule(lr){3-4}
\textbf{Type} & \textbf{Range} & \textbf{Type} & \textbf{Range} \\
\midrule
Brightness   & $(0.8,\,1.2)$   & Perspective & $0.025$ \\
Contrast     & $(0.8,\,1.2)$   & Rotation    & $[-5^{\circ},\,5^{\circ}]$ \\
Saturation   & $(0.5,\,1.5)$   & Crop        & $[10,\,40]$\,px \\
Hue          & $(-0.05,\,0.05)$ & & \\
Sharpness    & $(0.5,\,1.5)$   & & \\
\bottomrule
\end{tabular}
\caption{\textbf{Image augmentation configuration.} For color transformations, numeric ranges denote multiplicative or additive jitter factors applied to image intensities. For spatial transformations, ranges specify the perturbation magnitudes for projective distortion, rotation, and cropping.}
\label{tab:img-aug}
\vspace{-10pt}
\end{table}

\begin{table}[t]
\centering
\scriptsize
\setlength{\tabcolsep}{4pt}
\begin{tabular}{lccccc}
\toprule
\textbf{Model} & \textbf{Visual Res.} & \textbf{State Dim.} & \textbf{Action Dim.} & $\mathbf{T_p}$ & $\mathbf{T_e}$ \\
\midrule
ACT     & L:\,$120{\times}212$;\,H:\,$240{\times}240$ & 8 & 8 & 50 & 50  \\
DP      & L:\,$120{\times}212$;\,H:\,$240{\times}240$ & 8 & 8 & 64 & 50  \\
SmolVLA & L:\,$120{\times}212$;\,H:\,$240{\times}240$ & 8 & 8 & 50 & 50  \\
Pi-0    & L:\,$120{\times}212$;\,H:\,$240{\times}240$ & 8 & 8 & 50 & 50  \\
\bottomrule
\end{tabular}
\caption{\textbf{Observation and action spaces.} 
Low-resolution inputs are used for the rope-routing task, while high-resolution inputs are used for the other tasks. 
State and action vectors include end-effector position, quaternion, and gripper state, expressed in either absolute or relative coordinates. 
$\mathbf{T_p}$ and $\mathbf{T_e}$ denote the prediction and execution horizons, respectively.}
\label{tab:spaces}
\end{table}

\begin{table}[t]
\centering
\scriptsize
\setlength{\tabcolsep}{5pt}
\begin{tabular}{lccccc}
\toprule
\textbf{Vision Backbone} & \textbf{\#V-Params} & \textbf{\#P-Params} & \textbf{LR} & \textbf{Batch Size} & \textbf{\#Iters} \\
\midrule
ResNet-18 (ACT)   & 18M   & 34M   & $1{\times}10^{-5}$ & 512 & 7k  \\
ResNet-18 (DP)   & 18M   & 245M  & $1{\times}10^{-4}$ & 512 & 7k  \\
SmolVLM-2   & 350M  & 100M  & $1{\times}10^{-4}$ & 128 & 20k \\
PaliGemma (Pi-0)   & 260B  & 300M  & $5{\times}10^{-5}$ & 8   & 30k \\
\bottomrule
\end{tabular}
\caption{\textbf{Training configuration.} 
Model-specific hyperparameters used in policy training. 
$\text{\#V-Params}$ and $\text{\#P-Params}$ denote the number of parameters in the visual encoder and policy head, respectively. 
\textbf{LR}, \textbf{Batch Size}, and \textbf{\#Iters} refer to the learning rate, batch size, and total training iterations.}
\label{tab:training-config}
\vspace{-10pt}
\end{table}

\subsection{Policy Training}
\subsubsection{Datasets}
To better understand the data distribution used for both policy training and evaluation, we visualize the coverage of initial states in Figure~\ref{fig:coverage}.

\subsubsection{Normalizations}
Normalization plays a crucial role in ensuring stable policy learning and consistent performance across models. For input and output normalization, we follow the conventions defined in each algorithm’s original implementation (summarized in Table~\ref{tab:normalization}). Specifically, the \textit{mean–std} scheme standardizes features to zero mean and unit variance, whereas the \textit{min–max} scheme scales each dimension independently to $[-1,1]$. 

For the VLA (SmolVLA and Pi-0) policies, we employ \textit{relative actions} to encourage more corrective and stable behavior, treating each action as an $\text{SE}(3)$ transformation of the end-effector pose in the base frame. Inspired by \cite{trilbmteam2025careful}, we compute both normalization statistics (\textit{mean–std} or \textit{min–max}) over a rolling window corresponding to the action chunk size across the entire dataset. Each action within a chunk is then normalized using its own statistics to maintain a consistent magnitude in the normalized space—mitigating the tendency of later actions in the chunk to exhibit larger amplitudes.

\subsubsection{Image Augmentations}
To improve visual robustness and generalization, we apply a combination of color and spatial augmentations to each input image during training. 
For every image in a training batch, three augmentation operations are randomly sampled and composed. 
Table~\ref{tab:img-aug} summarizes the augmentation types and their corresponding parameter ranges.

\subsubsection{Hyperparameters}
A complete overview of the observation and action spaces, as well as the training configurations for each model, is presented in Tables~\ref{tab:spaces} and~\ref{tab:training-config}. 
For VLA-based policies, we finetune only the action head (keeping the pretrained vision-language encoder frozen) on our datasets.

\subsection{Evaluation}
\subsubsection{Evaluation Protocol}
During evaluation, we sample a fixed set of initial states, and rollout the policies from both sim and real. To ensure that sim and real align with each other, we first sample object initial states in simulation and render them from the same camera viewpoint as the real-world physical setup. Then, we save the set of initial frame renderings, and a real-time visualizer overlays these simulated states onto the live camera stream, enabling a human operator to manually adjust the objects to match the simulated configuration. 

\subsubsection{Episode Settings} In all evaluation experiments in the main paper, the number of episodes for each task and the grid-based object initial configuration randomization ranges (relative to their prespecified poses) are set as in Table \ref{tab:randomization}. 

\begin{table}[t]
\centering
\setlength{\tabcolsep}{4pt}
\begin{tabular}{lcccc}
\toprule
\textbf{Task} & \textbf{Episodes} & \textbf{$x$ (cm)} & \textbf{$y$ (cm)} & \textbf{$\theta$ (deg)} \\ 
\midrule
Toy packing (toy) & 20 & $[-5, 5]$ & $[-5, 3]$ & $[-5, 5]$ \\
Toy packing (box)   & 20 & $[-5, 5]$ & $[0, 5]$  & $[-5, 5]$ \\
Rope routing (rope) & 27 & $[-5, 5]$ & $[-5, 5]$ & $[-10, 10]$ \\
T-block pushing (T-block)     & 16 & $[-5, 5]$ & $[-5, 5]$ & $\{\pm 45, \pm 135\}$ \\
\bottomrule
\end{tabular}
\caption{\textbf{Task randomization ranges used for evaluation.} For each task, the initial object configurations are randomized: the plush toy and box in toy packing, the rope in rope routing, and the T-block in T-block pushing.}
\label{tab:randomization}
\vspace{-15pt}
\end{table}

\subsubsection{Success Criteria} Real robot experiments typically rely on human operators to record success and failure counts, which is tedious and introduces human bias. For simulated experiments to scale up, automated success criteria are necessary. For all three tasks, we design metrics based on simulation states as follows:

\paragraph{Toy Packing} For each frame, we calculate the number of PhysTwin mass particles that fall within an oriented bounding box of the box's mesh. Within the final 100 frames (3.3 seconds) of a 15-second episode, if the number exceeds a certain threshold for over 30 frames, the episode is considered successful. Empirically, the total number of PhysTwin points is 3095, and we use a threshold number of 3050.

\paragraph{Rope Rouing} For each frame, we calculate the number of PhysTwin spring segments that pass through the openings of the channel of the clip. Within the final 100 frames (3.3 seconds) of a 30-second episode, if for both openings and more than 30 frames, the number of the spring segments that cross the opening is over 100, that indicates a sufficient routing through the clip and the episode is considered successful.

\paragraph{T-block Pushing} For each frame, we calculate the mean squared Euclidean distance between the current PhysTwin particles and the target-state PhysTwin particles. Within the final 100 frames (3.3 seconds) of a 60-second episode, if the mean squared distance is less than 0.002, the episode is considered successful.

\begin{figure*}[t]
    \centering
    \includegraphics[width=0.75\textwidth]{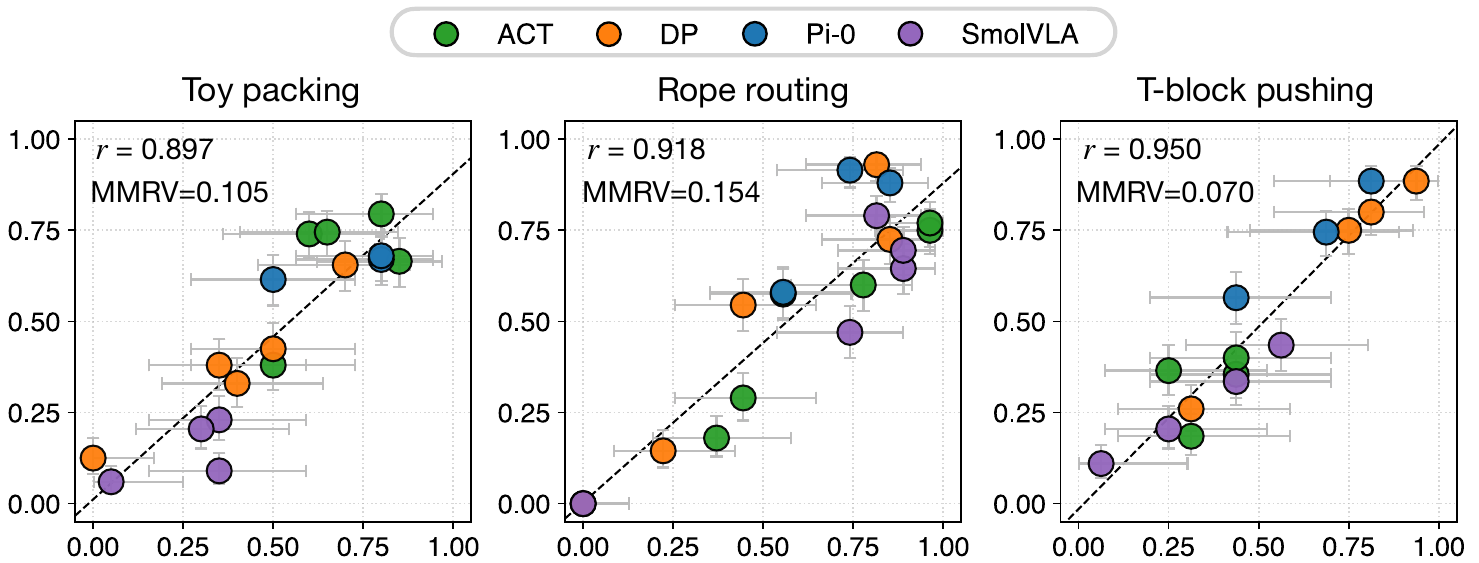}
    \caption{\textbf{Sim-and-real correlations from scaled-up simulation evaluations.} Each point represents a policy evaluated on both domains, and the shaded region indicates the 95\% confidence interval. Increasing the number of simulated episodes reduces statistical uncertainty and yields stable correlation estimates with real-world success rates, with the minimum observed correlation coefficient of 0.897. Compared to the main-paper experiments, the relative ordering of policy checkpoints remains consistent, demonstrating the robustness of the evaluation across larger-scale simulations.}
    \label{fig:corr-200}
    \vspace{-15pt}
\end{figure*}

\section{Additional Results}
\label{appendix:results}

\subsection{Scaling up Simulation Evaluation}
In the main paper, we evaluate each policy in simulation using an identical set of initial states as in the real-world experiments. This design controls for randomness but limits the number of available trials and thus results in high statistical uncertainty, as reflected by the wide Clopper-Pearson confidence intervals.

To account for the distributional differences introduced by uniformly sampling within the randomization range, we adopt slightly modified randomization settings compared to the grid-range experiments in the main paper. In the toy packing task, we use the same randomization range as described previously. For the rope routing task, we enlarge the $x, y, \theta$ randomization ranges to $[-7.5, 7.5]$ cm and $[-15, 15]$ degrees, respectively. For the T-block pushing task, we enlarge the $x$ and $y$ range to $[-7.5, 7.5]$ cm.

To better estimate the asymptotic correlation between simulation and real-world performance, we further scale up the number of simulation evaluations by sampling 200 randomized initial states from the task distribution. Figure~\ref{fig:corr-200} reports the resulting correlations between the scaled-up simulation metrics and real-world success rates. 

We observe that the confidence intervals are significantly narrowed down, and the correlation estimates stabilize as the number of simulation episodes increases, suggesting that simulation fidelity becomes a reliable predictor of real-world outcomes when averaged across diverse task instances.

\subsection{Replaying Real Rollouts}

To further assess correspondence between our simulation and the real world, we perform replay-based evaluations, where real-world rollouts during policy inference are re-executed in the simulator using the same control commands. This allows us to disentangle dynamic discrepancies from appearance gaps, i.e., the difference in policy behaviors introduced by differences in perceived images is eliminated. 

In total, we replay the real-world rollouts of 16 checkpoints each with 20 episodes for toy packing, 15 checkpoints each with 27 episodes for rope routing, and 12 checkpoints each with 16 episodes for T-block pushing. The object states in simulation are initialized to be identical to the corresponding real episodes.

\begin{figure*}[t]
    \centering
    \includegraphics[width=0.75\textwidth]{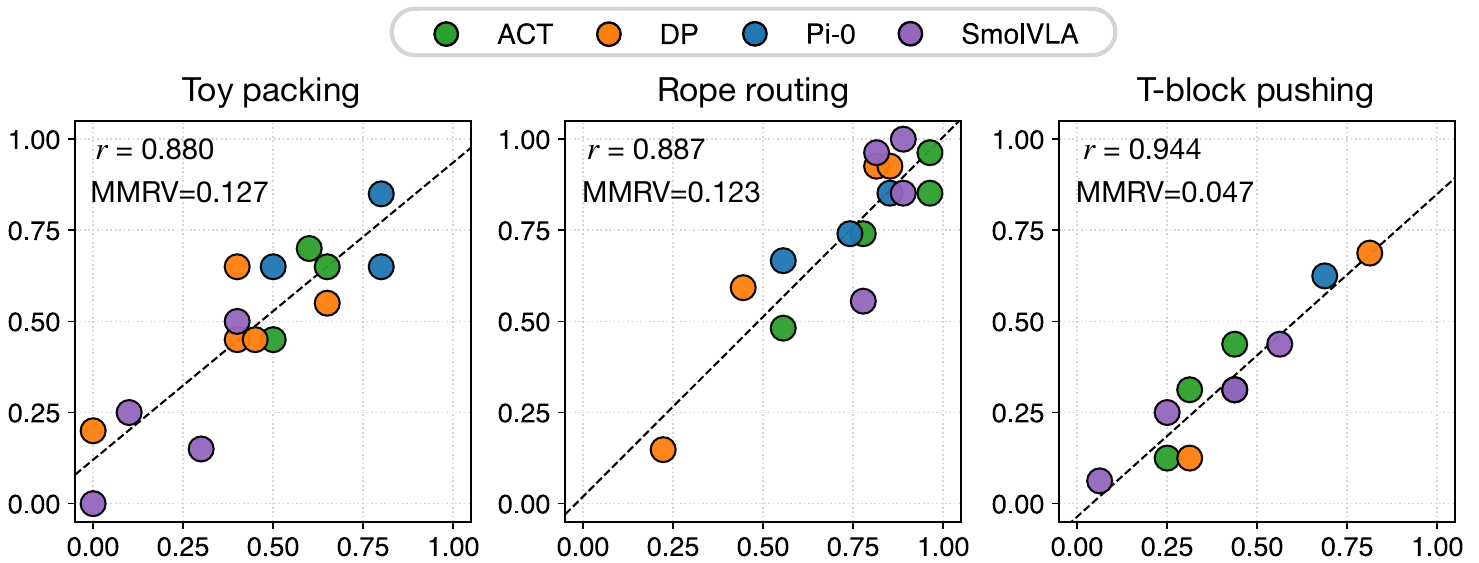}
    \caption{\textbf{Sim-and-real correlations from replaying real-world rollouts.} Each point corresponds to a replay of a real-world policy checkpoint's evaluation results using identical control commands and camera trajectories within the simulator. The success rates are averaged over all episodes for each checkpoint. The resulting alignment highlights the degree to which our simulator reproduces the observed real-world outcomes.}
    \label{fig:corr-replay}
\end{figure*}

Figure~\ref{fig:corr-replay} shows the resulting correlations, and Table~\ref{tab:confusion} reports the per-episode replay statistics. Across all three tasks, the confusion matrices exhibit strong diagonal dominance, indicating high agreement between replayed and real outcomes.

Notably, for toy packing, false positives (replayed success but real failure) are more frequent than false negatives, reflecting that the simulator tends to slightly overestimate success, likely due to simplified contact or friction models. For T-block pushing, false negatives are more frequent than false positives, indicating that some real success trajectories cannot be reproduced in the simulation, potentially due to a slight mismatch in friction coefficient and initial states. 

Overall, the high diagonal values highlight that the simulator can reproduce real rollout outcomes most of the time, even with pure open-loop trajectory replay. 

\begin{table*}[t]
\centering
\footnotesize
\setlength{\tabcolsep}{6pt}
\begin{tabular}{ccc}
\toprule
\textbf{Toy packing} & \textbf{Rope routing} & \textbf{T-block pushing} \\
\midrule
\begin{tabular}{c|cc}
 & GT\,+ & GT\,-- \\ \midrule
Replay\,+ & 106 & 37 \\
Replay\,-- & 25 & 132
\end{tabular}
&
\begin{tabular}{c|cc}
 & GT\,+ & GT\,-- \\ \midrule
Replay\,+ & 276 & 28 \\
Replay\,-- & 24 & 77
\end{tabular}
&
\begin{tabular}{c|cc}
 & GT\,+ & GT\,-- \\ \midrule
Replay\,+ & 63 & 1 \\
Replay\,-- & 17 & 111
\end{tabular} \\
\bottomrule
\end{tabular}
\caption{\textbf{Per-episode replay result.} We calculate the per-episode correlation between the replayed result and the real-world ground truth. Each subtable shows a $2\times2$ confusion matrix for each task (\textbf{TP}, \textbf{FP}, \textbf{FN}, \textbf{TN}), where rows indicate replay outcomes and columns indicate ground truth. Each entry records the total number of episodes, summed across all policy checkpoints. The strong diagonal dominance reflects high sim–real agreement in replayed trajectories.}
\label{tab:confusion}
\vspace{-10pt}
\end{table*}

\subsection{Additional Qualitative Results}
\begin{figure*}[t]
    \centering
    \includegraphics[width=\textwidth]{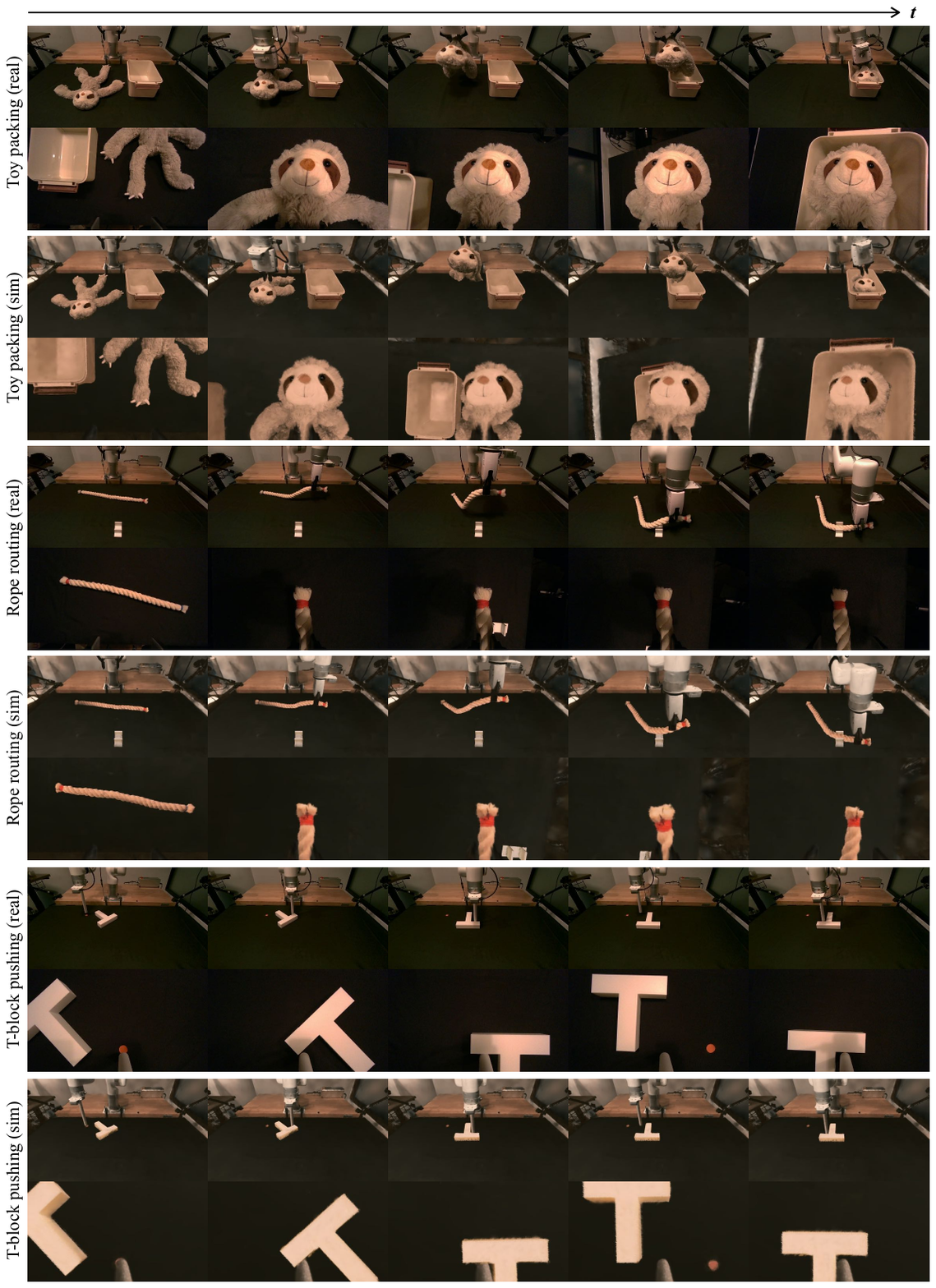}
    \caption{\textbf{Sim and real rollout trajectories.} Columns correspond to synchronized timesteps along each rollout, with identical timestamps selected for simulation and real-world policy rollouts to illustrate correspondence. Each panel (e.g., toy packing (real)) shows front-view (top) and wrist-view (bottom) observations, with panels alternating between real and simulated trajectories.}
    \label{fig:rollout_collage}
\end{figure*}

We include further visualizations in Figure~\ref{fig:rollout_collage}, which compares synchronized simulation and real-world trajectories across representative timesteps. For each task, we display both front and wrist camera views.

From the figure, we observe that the simulated trajectories closely reproduce the real-world sequences in both front-view and wrist-view observations. Object poses, contact transitions, and end-effector motions remain consistent across corresponding timesteps, indicating that the simulator effectively captures the underlying task dynamics as well as visual appearance.

\newpage
